\documentclass{article}

\usepackage{microtype}
\usepackage{graphicx}
\usepackage{booktabs} 

\usepackage{hyperref}

\usepackage{subcaption}

\usepackage[accepted]{icml2020}

\icmltitlerunning{Wiki-CS: A Wikipedia-Based Benchmark for Graph Neural Networks}

\begin{document}

\twocolumn[
\icmltitle{Wiki-CS: A Wikipedia-Based Benchmark for Graph Neural Networks}



\icmlsetsymbol{equal}{*}

\begin{icmlauthorlist}
\icmlauthor{Peter Mernyei}{cam}
\icmlauthor{C\u at\u alina Cangea}{cam}
\end{icmlauthorlist}

\icmlaffiliation{cam}{Department of Computer Science, University of Cambridge, Cambridge, United Kingdom}

\icmlcorrespondingauthor{Peter Mernyei}{pmernyei@gmail.com}

\icmlkeywords{Machine Learning, ICML}

\vskip 0.3in
]



\printAffiliationsAndNotice{}  

\begin{abstract}
We present \textsc{Wiki-CS}, a novel dataset derived from Wikipedia for benchmarking Graph Neural Networks. The dataset consists of nodes corresponding to Computer Science articles, with edges based on hyperlinks and 10 classes representing different branches of the field. We use the dataset to evaluate semi-supervised node classification and single-relation link prediction models. Our experiments show that these methods perform well on a new domain, with structural properties different from earlier benchmarks. The dataset is publicly available, along with the implementation of the data pipeline and the benchmark experiments, at \url{https://github.com/pmernyei/wiki-cs-dataset}.
\end{abstract}

\section{Introduction}
In recent years, significant advances have been made in learning representations of graph-structured data and predicting quantities of interest for nodes, edges or graphs themselves~\cite{gcn,gat}. This new subfield has attracted an increasing amount of interest, leading to the development of numerous methods~\cite{2019survey}. However, several earlier works have noted issues with existing standard benchmarks, which make it difficult to rigorously compare results and accurately distinguish between the performance of competing architectures~\cite{oleks2018pitfalls,appnp}.

Our primary focus is semi-supervised node classification: given labels of a small subset of nodes (typically 1--5\%) and features of all nodes, as well as their connectivity information, the task is to predict all other labels. This setup is often used to assess the performance of various Graph Neural Network (GNN) architectures~\cite{gcn,gat}. These methods are usually evaluated on three citation network datasets (Cora, CiteSeer, PubMed) introduced by Yang et al~\yrcite{planetoid}. Unfortunately, different training splits are used across studies, which makes comparisons challenging, especially since the performance on this task is very sensitive to the choice of training split~\cite{oleks2018pitfalls}. Furthermore, all three benchmarks are derived from the same domain, with similar structural properties. In contrast, \textsc{Wiki-CS} has a much higher connectivity rate, hence it provides a different kind of distribution for these methods to be tested on.

Similar datasets have also been used in single-relation link prediction~\cite{vgae,graphstar}. We further use \textsc{Wiki-CS} to benchmark relational methods for this task, along with a non-structural SVM baseline. 

\section{Related Work}
The most commonly used semi-supervised node classification benchmarks are the previously-described citation network graphs, proposed by Yang et al.~\yrcite{planetoid}. Larger datasets have also been used, such as Reddit and PPI~\cite{graphsage}. However, due to the standard split sizes proposed for these benchmarks, state-of-the-art methods have already achieved F1 scores of 0.995 and 0.97 respectively~\cite{graphsaint}, making it difficult for further improvements to be properly gauged.

Due to the issues with existing datasets, there has been significant concurrent work on establishing robust GNN benchmarks:
\begin{itemize}
    \item The Open Graph Benchmark~\cite{open-graph-benchmark} (OGB) has recently developed a range of datasets, focusing on diversity of domains, graph sizes and types of tasks and unified evaluation methods. A Wikidata Knowledge Graph is included for a link prediction task---note that this source material is entirely different from the article hyperlink graph used for \textsc{Wiki-CS}. OGB also proposes challenging domain-specific splits based on some aspect of the data (for exaxmple, time or molecular structure), instead of selecting this randomly.
    \item Dwivedi et al.~\yrcite{dwivedi2020benchmarking} similarly proposed several datasets to rigorously distinguish the aspects of GNN architectures that significantly contribute to good performance on challenging benchmarks. To achieve this, they used largely synthetic graphs.
\end{itemize}  
Our contribution complements the existing ones by providing a dataset and experimental results based on a new domain. We thus further establish the generality of GNN methods and extend the range of available benchmarks.

\section{The Dataset}

\subsection{Article Selection and Label Generation}
We processed Wikipedia datadumps from August 2019 to extract a subgraph where accurate class labels could be provided, based on the categories of each article. Unfortunately, these category tags were not suitable for direct use as class labels, as most of them are highly specific and inconsistently applied to a small number of pages---there were around 1.5 million different categories defined on the 6 million pages, at the time of the snapshot that we used.

\begin{figure}[ht]
\vskip 0.2in
\centering

\includegraphics[width=0.9\columnwidth]{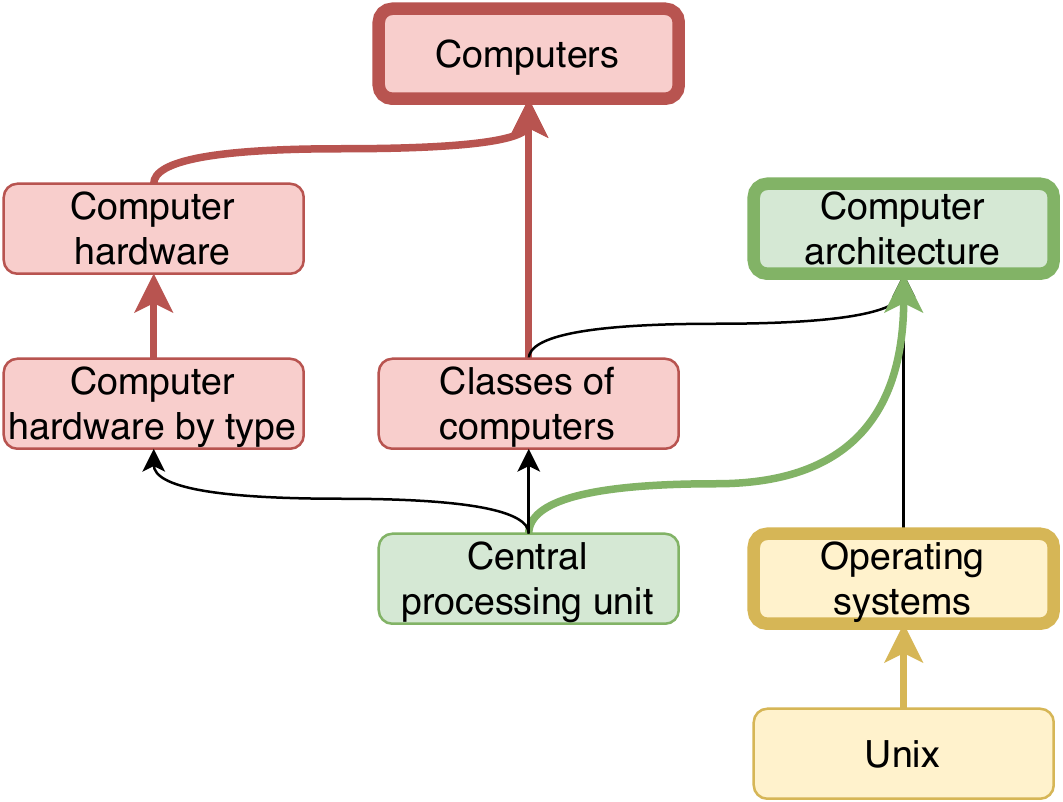} 

\caption{A subgraph of the subcategory relation graph. Nodes with dark borders are the prominent categories chosen based on centrality. The others were aggregated to the nearest marked ancestor as denoted by their colors, with ties broken arbitrarily.}

\label{fig:cs-categories}
\vskip -0.2in
\end{figure}

This problem was mitigated by using the category sanitizer tool made available by Boldi \& Monti~\yrcite{category-sanitizer}, with some modifications. Their method relies on the subcategory relation to aggregate articles belonging to subcategories to their parent categories. A small set of prominent categories is selected based on harmonic centrality measures in the subcategory graph; other nodes in the subcategory graph are aggregated to one of their nearest ancestors (see Figure \ref{fig:cs-categories} for an example subgraph). See Boldi \& Monti~\yrcite{category-sanitizer} for the details of the process. This avoids aggregation through many indirect steps, which would often lead to articles being mapped to categories which they have little semantic overlap with.

However, the output still required further clean-up: some aggregated categories still contained unrelated articles. Additionally, if the subcategory graph related to some topic is very dense, the selected prominent categories and the aggregation choices can be very arbitrary.

\textsc{Wiki-CS} was created by inspecting the list of $10,000$ prominent categories selected by the sanitizer and picking a subject area with few such issues. We identified three possible candidate subjects (branches of biology, US states, branches of CS), and sampled 20 pages from every class of these candidates. Although all of these had some issues, we were able to clean up the CS data by dropping some categories and manually disabling aggregation across specific subcategories to prune bad pages from others. This resulted in a dataset with 10 classes corresponding to branches of computer science, with very high connectivity. See Appendix \ref{app:categories} for the set of prominent categories we used for each label. Finally, we dropped the articles that would have been mapped to multiple classes.

\subsection{Node Features}

Similarly to previous work~\cite{planetoid}, our node features were derived from the text of the corresponding articles. However, they were calculated as the average of pre-trained GloVe word embeddings~\cite{glove} instead of using binary bag-of-words vectors. This allowed us to encode rich features corresponding to a large vocabulary in relatively small 300-dimensional input vectors, which can be an advantage for training large models on a GPU.

\subsection{Training Splits}
\label{subsec:training-splits}

It has been shown that the choice of the training split can seriously affect model performance for semi-supervised node classification~\cite{oleks2018pitfalls}. Therefore, using multiple training splits can improve the robustness of a benchmark~\cite{appnp}. For this reason, we randomly selected 20 different training splits from the data that was not used for testing.

More specifically, we split the nodes in each class into two sets, 50\% for the test set and 50\% potentially visible. From the visible set, we generated 20 different splits of training, validation and early-stopping sets: 5\% of the nodes in each class were used for training in each split, 22.5\% were used to evaluate the early-stopping criterion, and 22.5\% were used as the validation set for hyperparameter tuning. We stored the resulting mask vectors with the rest of the dataset, so that they can be used consistently across all future work.

\subsection{Statistics and Structural Properties}

\begin{table}[t]
\caption{Comparison of key dataset statistics between \textsc{Wiki-CS} and standard citation network benchmarks. SP stands for shortest path length.}
\label{tab:dataset-statistics}
\vskip 0.15in
\begin{center}
\begin{small}
\begin{sc}
\begin{tabular}{l|ccc|c}
\toprule
                                & \bf{Cora} & \bf{CiteSeer} & \bf{PubMed} & \bf{Wiki-CS}\\
\midrule
         \bf{Classes}               &   7   &       6       &       3     &     10   \\
         \bf{Nodes}                 &  2708 &       3327    &   19717     &  11701   \\
         \bf{Edges}                 &  5429 &       4732    &   44338     & 216123   \\
         \bf{Features dim.}         &  1433 &       3703    &     500     &    300   \\
         \bf{Label rate}            & 3.6\% &       5.2\%   &   0.3\%     &    5\%   \\
         \bf{Mean degree}           &  4.00 &       2.84    &   4.50      &  36.94   \\
         \bf{\shortstack{Average SP}}  & 6.31  &       9.32    &       6.34  &    3.01    \\
\bottomrule
\end{tabular}
\end{sc}
\end{small}
\end{center}
\vskip -0.1in
\end{table}

Table \ref{tab:dataset-statistics} summarises the key statistics of the citation network and the \textsc{Wiki-CS} datasets. Note the significantly higher rate of connectivity compared to existing benchmarks and the short average distance between any two nodes. This suggests that progress could be made on the benchmark by designing more involved computations within the neighborhood of a node, rather than focusing on long-range connections. This makes \textsc{Wiki-CS} a useful and complementary addition to existing node classification datasets.

This connectivity also leads to more varied node neighborhoods: for each node, we calculated the proportion of neighbors that belong to the same class as the node itself, and plotted this distribution for \textsc{Wiki-CS} as well as the existing citation network benchmarks. The results shown in Figure \ref{fig:same-class-neighbors} show that the existing datasets have a large share of nodes in homogeneous neighborhoods, while \textsc{Wiki-CS} is significantly more varied.

\begin{figure}
\vskip 0.2in
\centering

\begin{subfigure}{0.49\columnwidth}
\includegraphics[width=\linewidth]{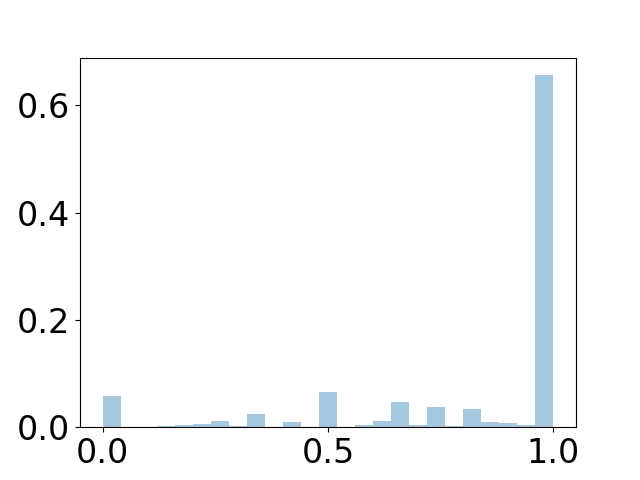} 
\caption{Cora}
\end{subfigure}
\begin{subfigure}{0.49\columnwidth}
\includegraphics[width=\linewidth]{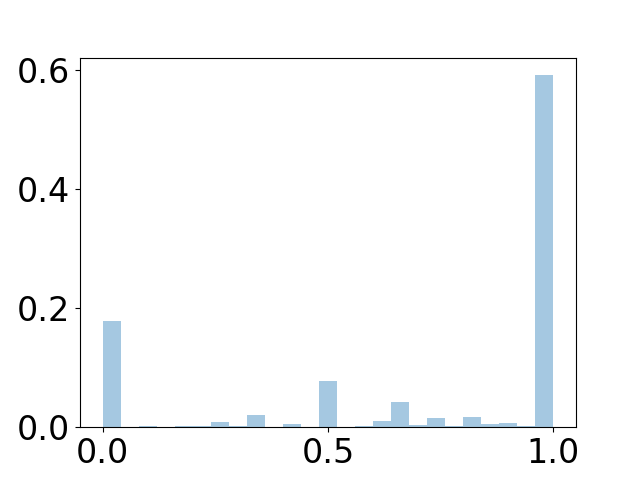} 
\caption{CiteSeer}
\end{subfigure}

\begin{subfigure}{0.49\columnwidth}
\includegraphics[width=\linewidth]{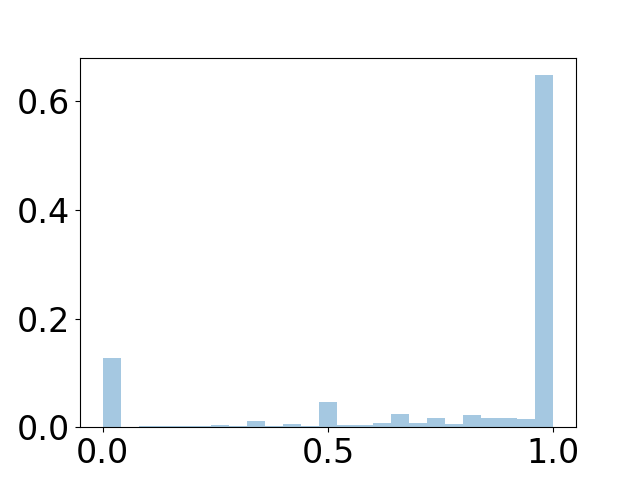} 
\caption{PubMed}
\end{subfigure}
\begin{subfigure}{0.49\columnwidth}
\includegraphics[width=\linewidth]{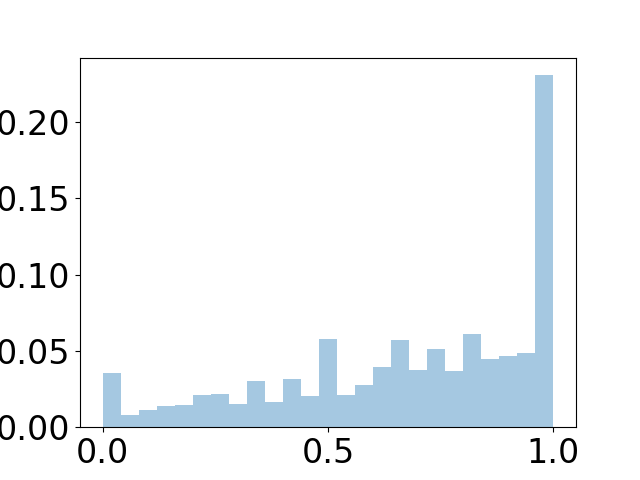} 
\caption{Wiki-CS}
\label{fig:same-class-neighbors}
\end{subfigure}

\caption{Distribution of the ratio of neighbors belonging to the same class. In all three the citation network datasets, almost two-thirds of all nodes have all neighbors belonging to the same class. The distribution of \textsc{Wiki-CS} is considerably more balanced.}
\label{fig:same-class-neighbors}
\end{figure}

\begin{figure}
\vskip 0.2in
\centering

\begin{subfigure}{0.49\columnwidth}
\includegraphics[width=\linewidth]{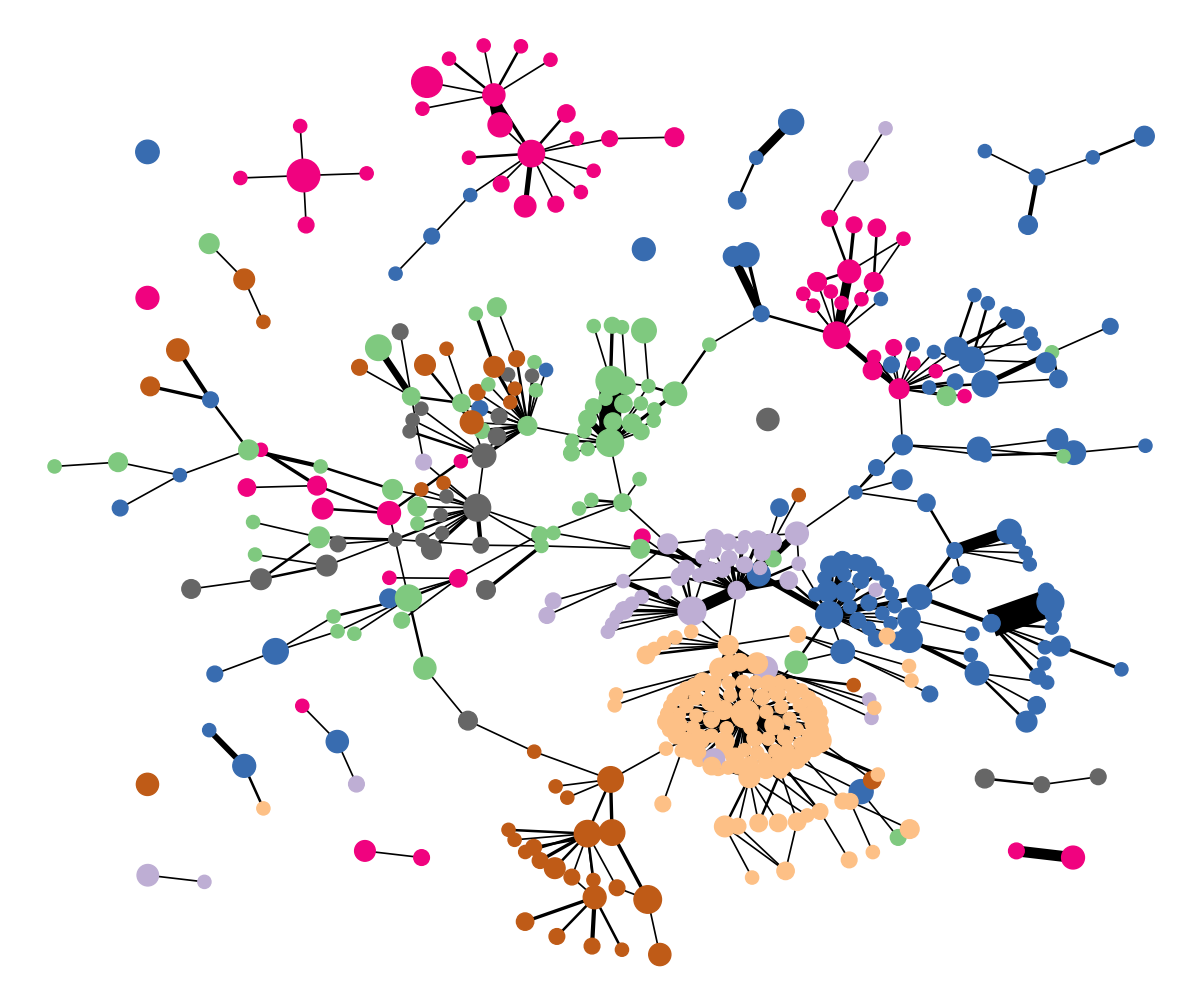} 
\caption{Cora}
\end{subfigure}
\begin{subfigure}{0.49\columnwidth}
\includegraphics[width=\linewidth]{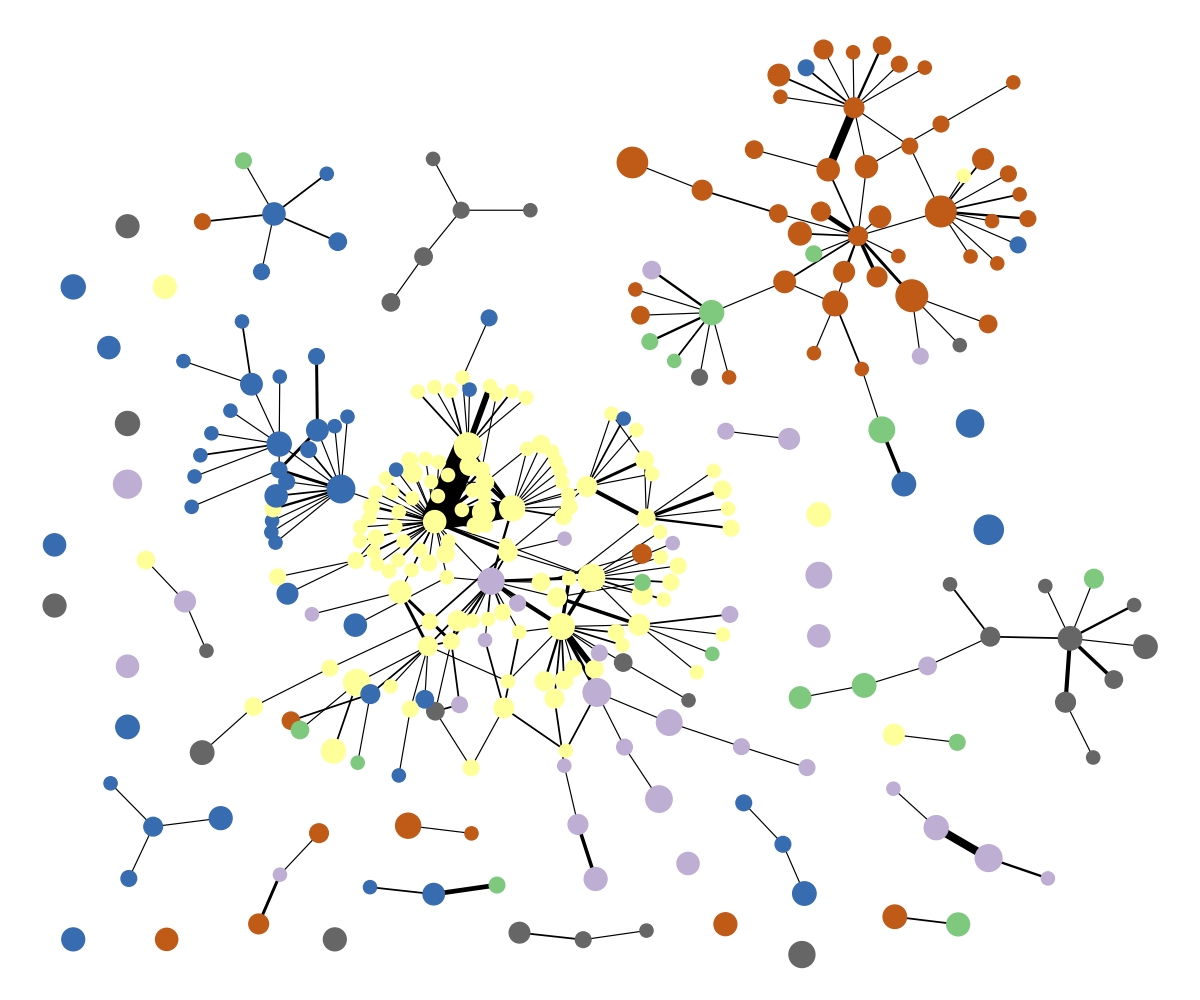} 
\caption{CiteSeer}
\end{subfigure}

\begin{subfigure}{0.49\columnwidth}
\includegraphics[width=\linewidth]{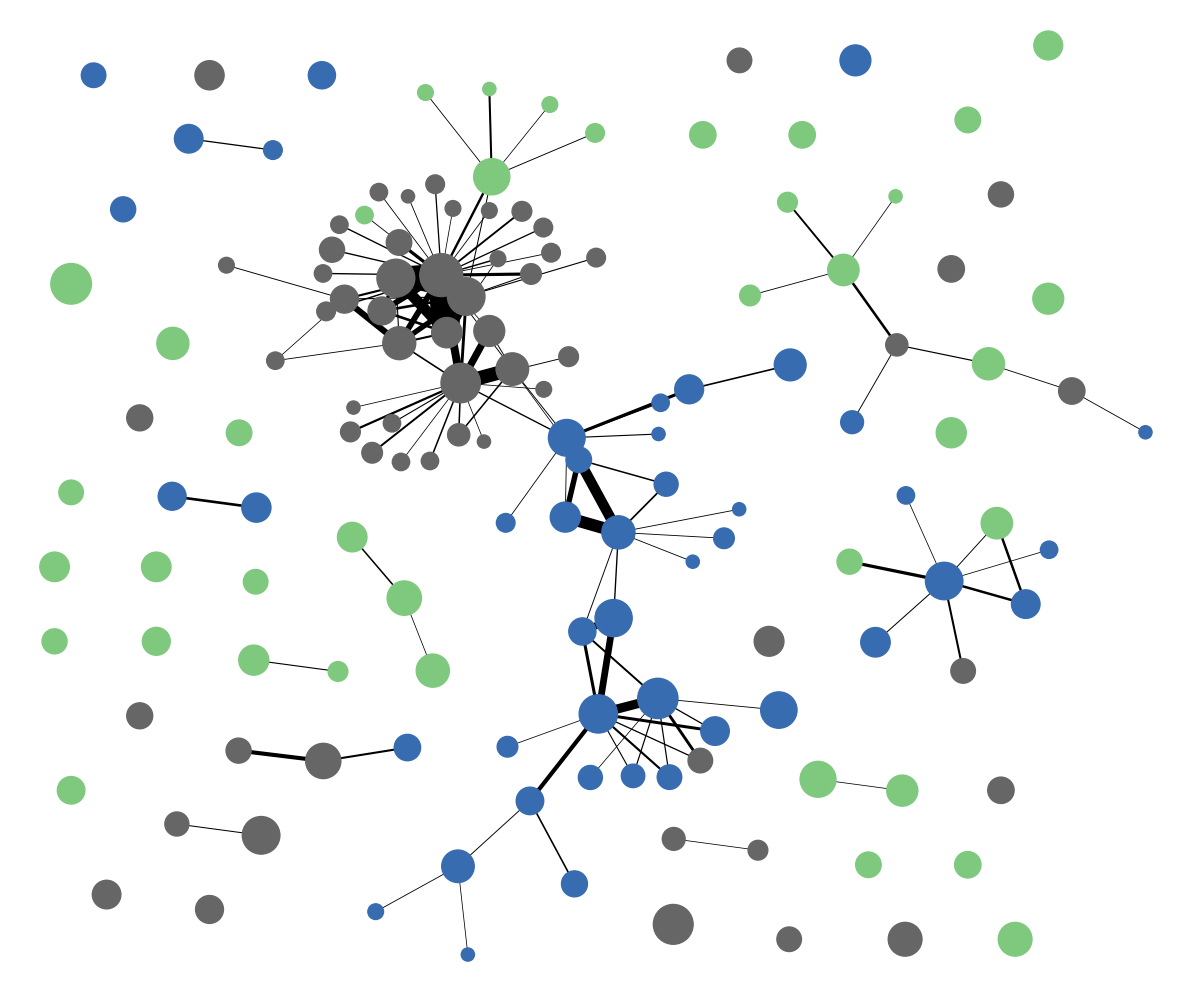} 
\caption{PubMed}
\end{subfigure}
\begin{subfigure}{0.49\columnwidth}
\includegraphics[width=\linewidth]{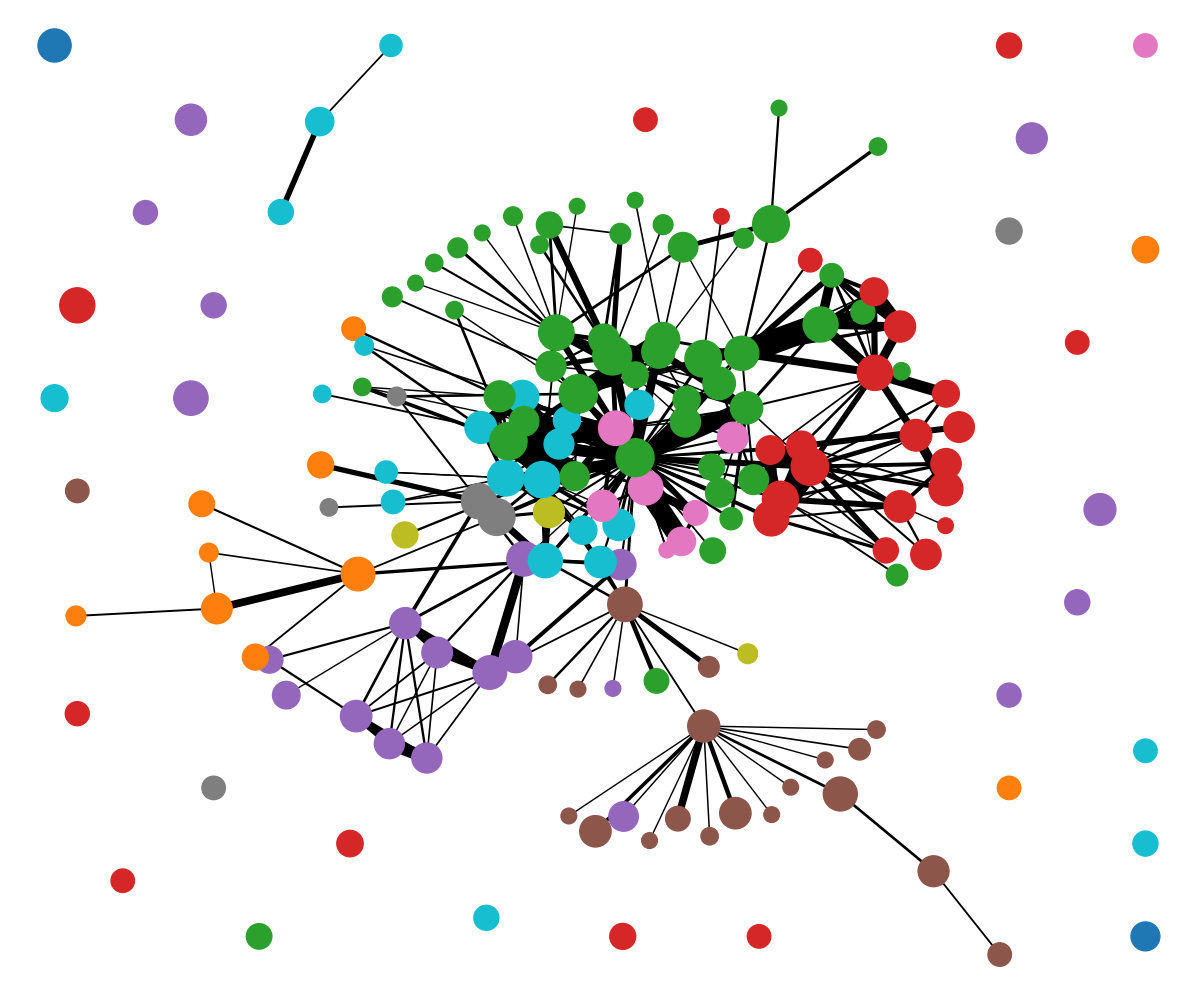} 
\caption{Wiki-CS}
\label{fig:dgm-vis-wiki-cs}
\end{subfigure}

\caption{Deep Graph Mapper (DGM) visualisation of benchmarks. Each node in the figure corresponds to a cluster of similar nodes in the original graph, with edge thickness representing the amount of connections between clusters. Colors represent the most frequent class in each cluster. The DGM unsupervised embedding process did not take labels into account, only relying on the node features and edges. The hyperparameters are described in Appendix \ref{app:hyperparameters}.}
\label{fig:dgm-vis}
\vskip -0.2in
\end{figure}

We also visualized the structure of all four datasets using Deep Graph Mapper~\cite{dgm}, an unsupervised GNN-based visualisation technique. The results shown in Figure \ref{fig:dgm-vis} suggest that \textsc{Wiki-CS} might have a more centralized, hierarchical structure than the citation networks, which seems plausible considering the different source domains.

\section{Experiments}

\subsection{Semi-Supervised Node Classification}

As described in Section \ref{subsec:training-splits}, 20 different training splits were created for the node classification task, each consisting of 5\% of nodes from each class. The same test set (50\% of the nodes) was evaluated for all splits. In each split, a different 22.5\% of nodes is used for early-stopping: we finish training when the loss calculated on this set has not improved for 100 epochs, and evaluate the model snapshot that produced the lowest loss.

This evaluation was performed 5 times on each of the 20 splits; we report the mean accuracy with a 95\% confidence interval based on bootstrap resampling from these results with 1,000 samples.

Three GNN models were evaluated: GCN~\cite{gcn}, GAT~\cite{gat} and APPNP~\cite{appnp}. Hyperparameter tuning was performed using the same training setup and measuring validation performance on 22.5\% of the nodes disjoint from the training and early-stopping sets. For efficiency, only the first 10 (out of 20) splits were used for hyperparameter tuning. The model configurations are described in Appendix \ref{app:hyperparameters}.

Two non-structural baselines were also included: a multi-layer perceptron (MLP) and a support vector machine (SVM). These predicted the class for each node individually, based on the node features. Since SVMs are deterministic, we only had a single data point from each training split and report the mean accuracy.

The results are shown in Table \ref{tab:node-classification}. The relative model performances align well with the results on citation network benchmarks, providing evidence that these are indeed good general-purpose methods. It is perhaps surprising that the attention mechanism of GAT improved very little on the GCN result despite the large neighborhoods---one reason might be that it is difficult to learn what to attend to in the semi-supervised setting, as discussed in-depth by Knyazev et al.~\yrcite{attention}.

The model predictions were also visualised with Deep Graph Mapper, and are included in Appendix \ref{app:dgm-preds}. This was based on training each model once, on the first of the 20 training splits. As expected, the mistakes and disagreements are largely located near boundaries of classes. This reinforces the idea that more complex neighborhood aggregation methods might be able to improve prediction accuracy. There are also some less connected clusters that seem to produce consistent incorrect predictions under all models---this might be due to not having good training samples in their proximity.

\begin{table}[ht]
\caption{Performance of semi-supervised node classification methods on the \textsc{Wiki-CS} dataset. Accuracies are represented as the average over 100 runs, with 95\% confidence intervals calculated by bootstrapping.}
\label{tab:node-classification}
\vskip 0.15in
\begin{center}
\begin{small}
\begin{sc}
    \begin{tabular}{c|c}
\toprule
          & \textbf{Accuracy} \\
\midrule
         \textbf{SVM} & 72.63\%\\
         \textbf{MLP} & 73.17 $\pm$ 0.19\%\\
\midrule
         \textbf{GCN} & 79.07 $\pm$ 0.10\%\\
         \textbf{GAT} & 79.63 $\pm$ 0.10\%\\
         \textbf{APPNP}&79.84 $\pm$ 0.10\%\\
\bottomrule
\end{tabular}
\end{sc}
\end{small}
\end{center}
\vskip -0.1in
\end{table}

\subsection{Link Prediction}

For the link prediction benchmark, we followed the experimental setup of studies performing single-relation link prediction on the Cora, CiteSeer and PubMed datasets~\cite{vgae, graphstar}. We split the data as follows: 85\% of the real edges for training, 5\% for validation and 10\% for testing. For each group, the same number of negative examples (non-edge node pairs) was sampled uniformly at random. 

Two GNN methods were benchmarked for link prediction: GraphStar~\cite{graphstar} and VGAE~\cite{vgae}. They were trained using the configurations reported in the original works, except for the hidden layer size of GraphStar: a maximum size of 256 would fit on the GPU. Details are included in Appendix \ref{app:hyperparameters}. An MLP baseline was also trained using concatenated pairs of node feature vectors. 

The results are shown in Table \ref{tab:link-prediction}. Note the extremely high performance of all models, even the MLP baseline. It appears that randomly selected false edges are very easy to distinguish from true edges in this dataset, and harder negative samples would be needed for more meaningful benchmarking. The large number of edges aggravates this, but it is not the main cause: we performed an experiment where we trained the models on just $10000$ examples of each class, and found the metrics to be still comfortably above $0.9$. See Table \ref{tab:lp-10k} for the results.we

\begin{table}[t]
\caption{Performance of link prediction methods on the \textsc{Wiki-CS} dataset. Metrics are represented as the average over 50 runs of VGAE, 20 runs of the MLP and 10 runs of GraphStar, with 95\% confidence intervals calculated by bootstrapping.}
\label{tab:link-prediction}
\vskip 0.15in
\begin{center}
\begin{small}
\begin{sc}
    \begin{tabular}{c|c c}
\toprule
          & \textbf{ROC-AUC} & \textbf{AP}  \\
\midrule
         \textbf{MLP} & $0.9785 \pm 0.0001$ & $0.9761 \pm 0.0002$ \\
\midrule
         \textbf{VGAE} & $0.9553 \pm 0.0008$ & $0.9608 \pm 0.0007$ \\
         \textbf{GraphStar} & $0.9793 \pm 0.0002$ & $0.9896 \pm 0.0001$\\
\bottomrule
\end{tabular}
\end{sc}
\end{small}
\end{center}
\vskip -0.1in
\end{table}

\begin{table}[t]
\caption{Performance of link prediction methods trained on only $10,000$ examples of each class.}
\label{tab:lp-10k}
\vskip 0.15in
\begin{center}
\begin{small}
\begin{sc}
    \begin{tabular}{c|c c}
\toprule
          & \textbf{ROC-AUC} & \textbf{AP}  \\
\midrule
         \textbf{MLP} & $0.9192 \pm 0.0004$ & $0.9119 \pm 0.0006$ \\
\midrule
         \textbf{VGAE} & $0.8546 \pm 0.0024$ & $0.8427 \pm 0.0032$ \\
         \textbf{GraphStar} & $0.9577 \pm 0.0006$ & $0.9795 \pm 0.0003$\\
\bottomrule
\end{tabular}
\end{sc}
\end{small}
\end{center}
\vskip -0.1in
\end{table}

\section{Conclusion}

We have presented \textsc{Wiki-CS}, a new benchmark for GNN methods. We have described how its structural properties are significantly different from commonly used datasets. Our experiments show existing GNN architectures for semi-supervised node classification and link prediction performing similarly to their results on other benchmarks, which is further evidence that they are good general-purpose methods for graph-learning tasks. Our dataset is available for further study, broadening the range of available benchmarks.

\bibliography{paper}
\bibliographystyle{icml2020}

\clearpage
\appendix
\section{Hyperparameter settings}
\label{app:hyperparameters}
\subsection{Node classification}
The following hyperaparameters were used for the node classification models:
\begin{itemize}
    \item \textbf{MLP:} 2 layers, $35$ units in the hidden layer, $0.35$ dropout probability, learning rate $0.003$.
    \item \textbf{SVM:} radial basis function kernel with $C=8$.
    \item \textbf{GCN:} 2 layers, 33 hidden neurons, learning rate $0.02$, $0.25$ dropout probability, with self-loops added at each node.
    \item \textbf{GAT:} 2 layers, 5 attention heads for the hidden layer outputting 14 units each, $0.5$ dropout probability, learning rate $0.007$, with self-loops added at each node.
    \item \textbf{APPNP:} learning rate $0.02$, $0.4$ dropout probability, propagation iterations $k=2$, teleport probability $\alpha=0.11$.
\end{itemize}

Additionally, all models used an L2 loss coefficient of $5\times10^{-4}$.

\subsection{Link prediction}
\begin{itemize}
    \item \textbf{GraphStar:} the original training code with a dynamically varying learning rate schedule was used, trained for 1,000 epochs and the test metrics reported from the epoch with the highest average validation metrics. A hidden layer size of 256 was used to fit into memory and all other parameters as given by the authors: 3 layers, no dropout, $5\times 10^{-4}$ L2 regularisation.
    \item \textbf{VGAE:} largely the same setup was used as the original paper: 16-dimensional latent space, GCN encoders with a shared hidden layer of 32 neurons, no dropout, trained for 200 epochs. However, we found improved performance when discarding the KL regularisation loss on the latent space distributions. 
    \item \textbf{MLP:} a fully connected network with 3 hidden layers of 128 units each, trained for 100 epochs with $0.2$ dropout probability after each layer, learning rate $10^{-4}$ and no weight decay.
\end{itemize}
\newpage
\subsection{DGM visualisations}
\begin{figure}[h]
\vskip 0.2in
\centering

\begin{subfigure}{0.49\columnwidth}
\includegraphics[width=\linewidth]{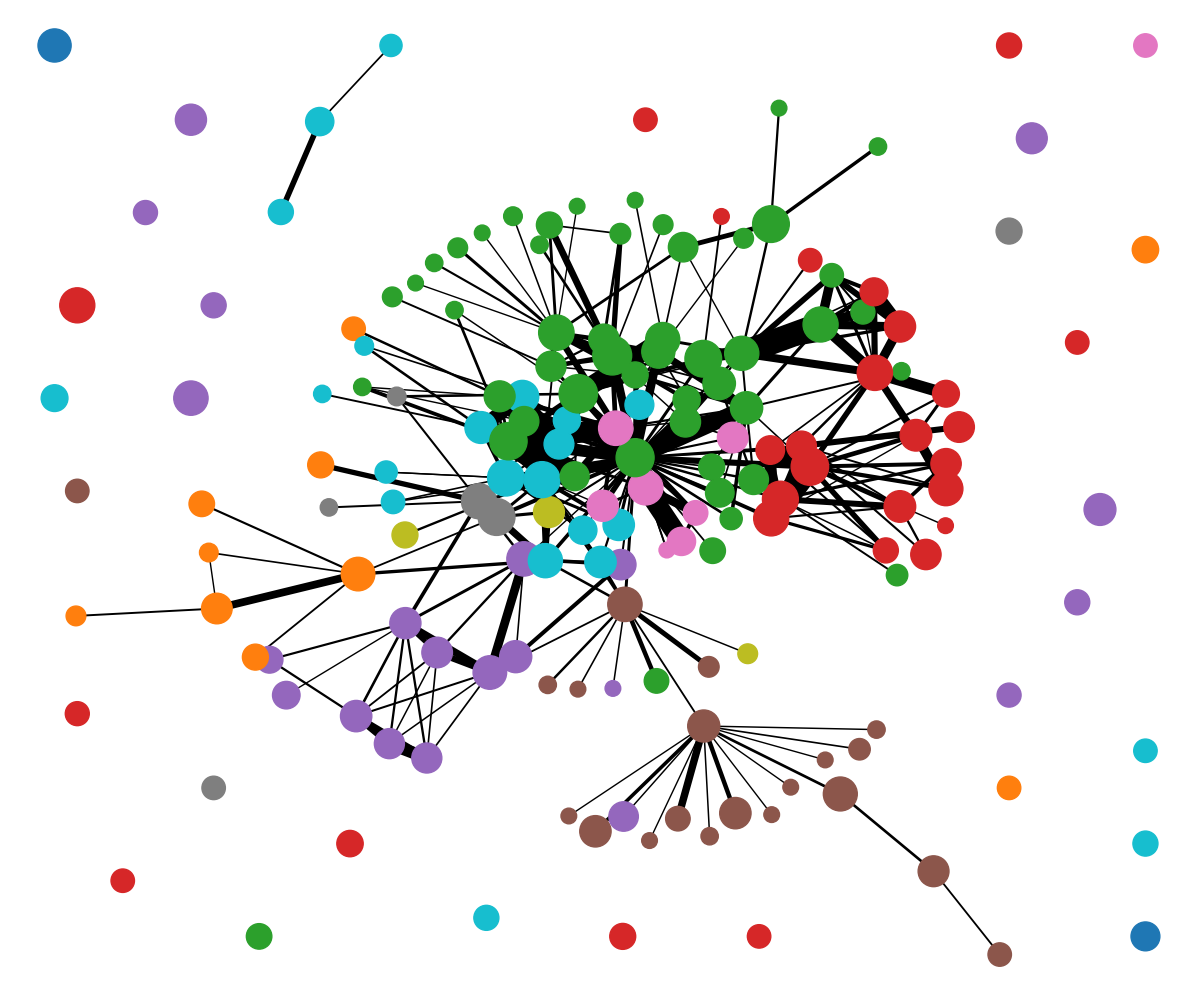} 
\caption{Ground truth labels}
\end{subfigure}

\begin{subfigure}{0.49\columnwidth}
\includegraphics[width=\linewidth]{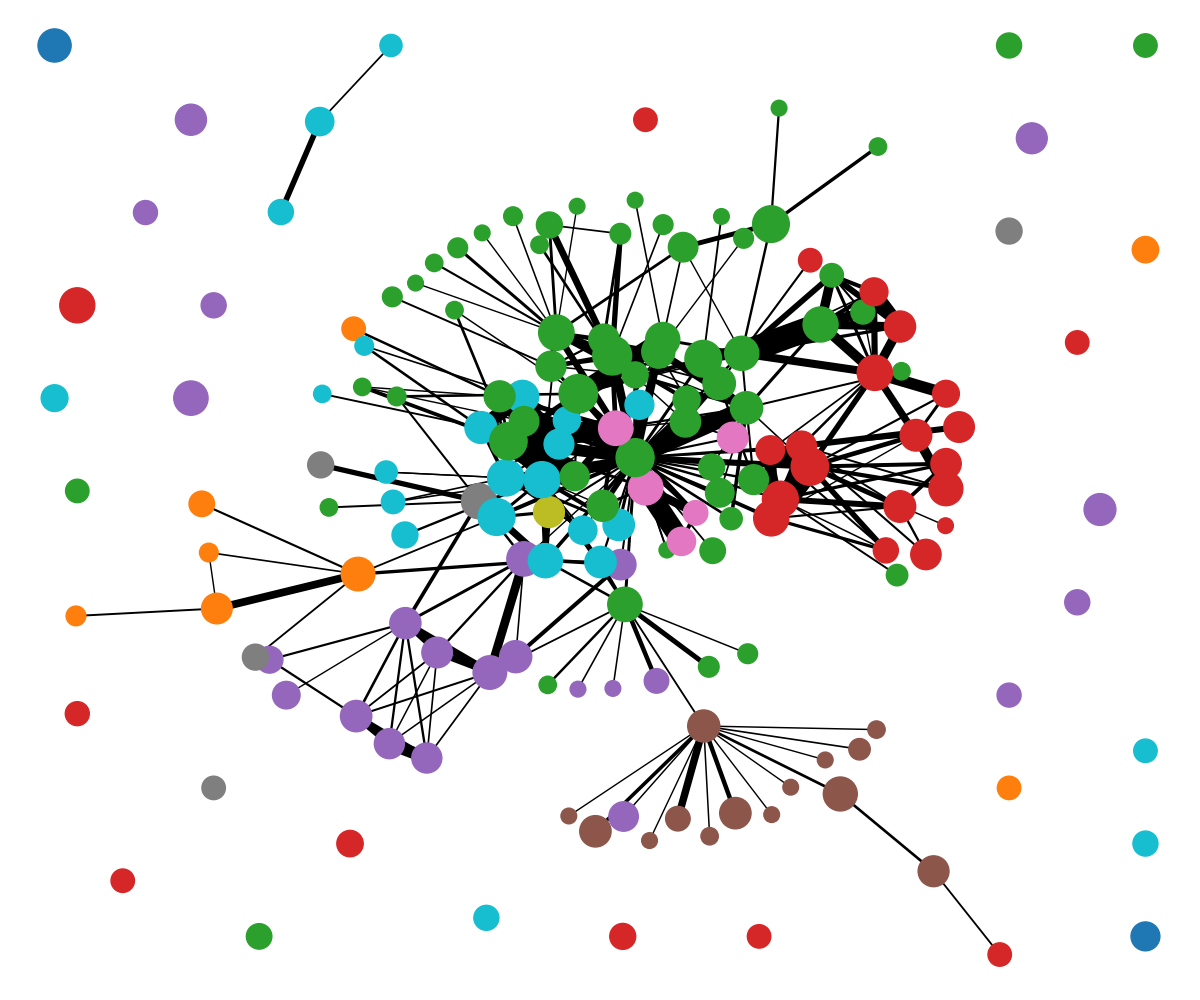} 
\caption{MLP predictions}
\end{subfigure}
\begin{subfigure}{0.49\columnwidth}
\includegraphics[width=\linewidth]{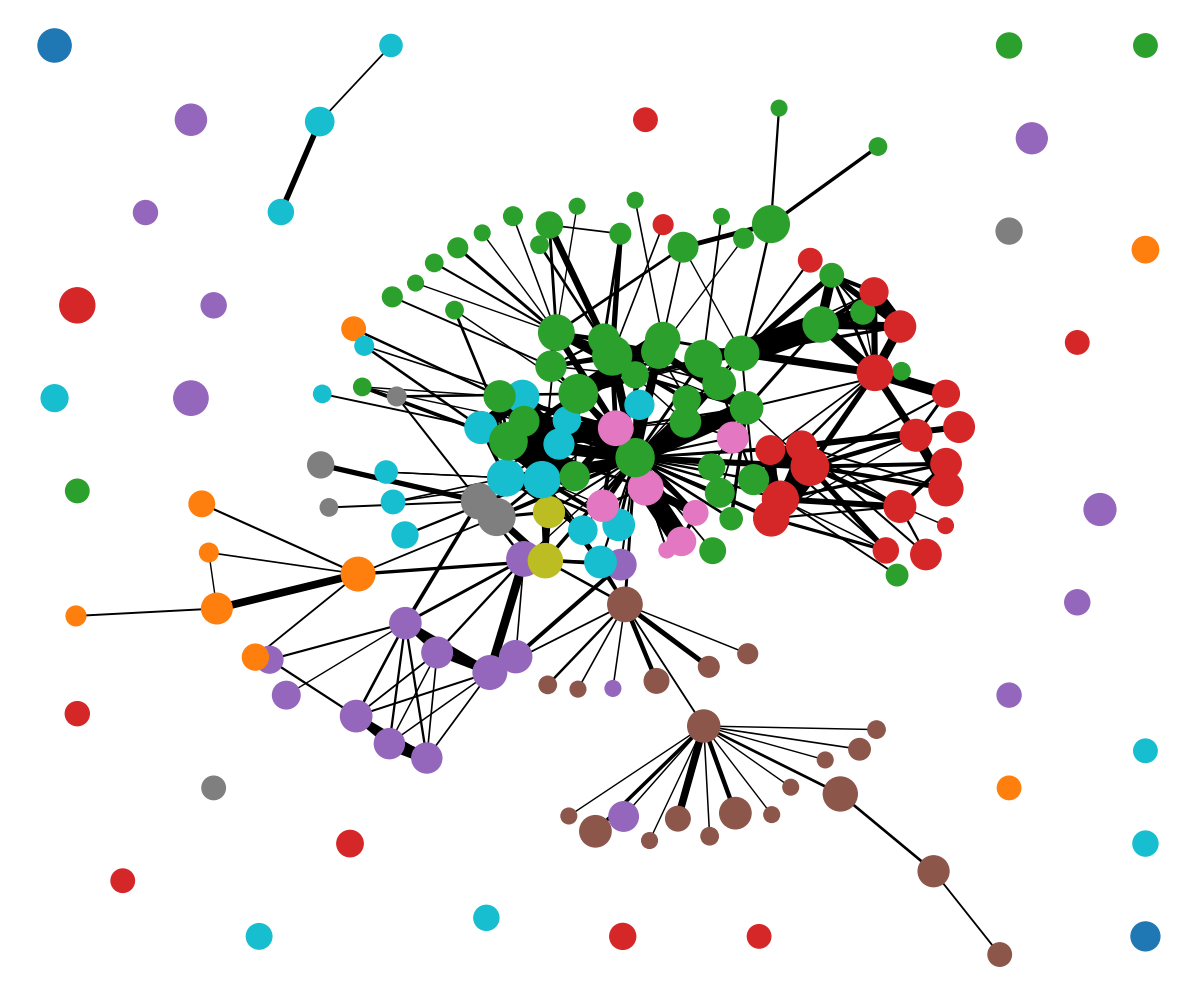} 
\caption{GCN predictions}
\end{subfigure}

\begin{subfigure}{0.49\columnwidth}
\includegraphics[width=\linewidth]{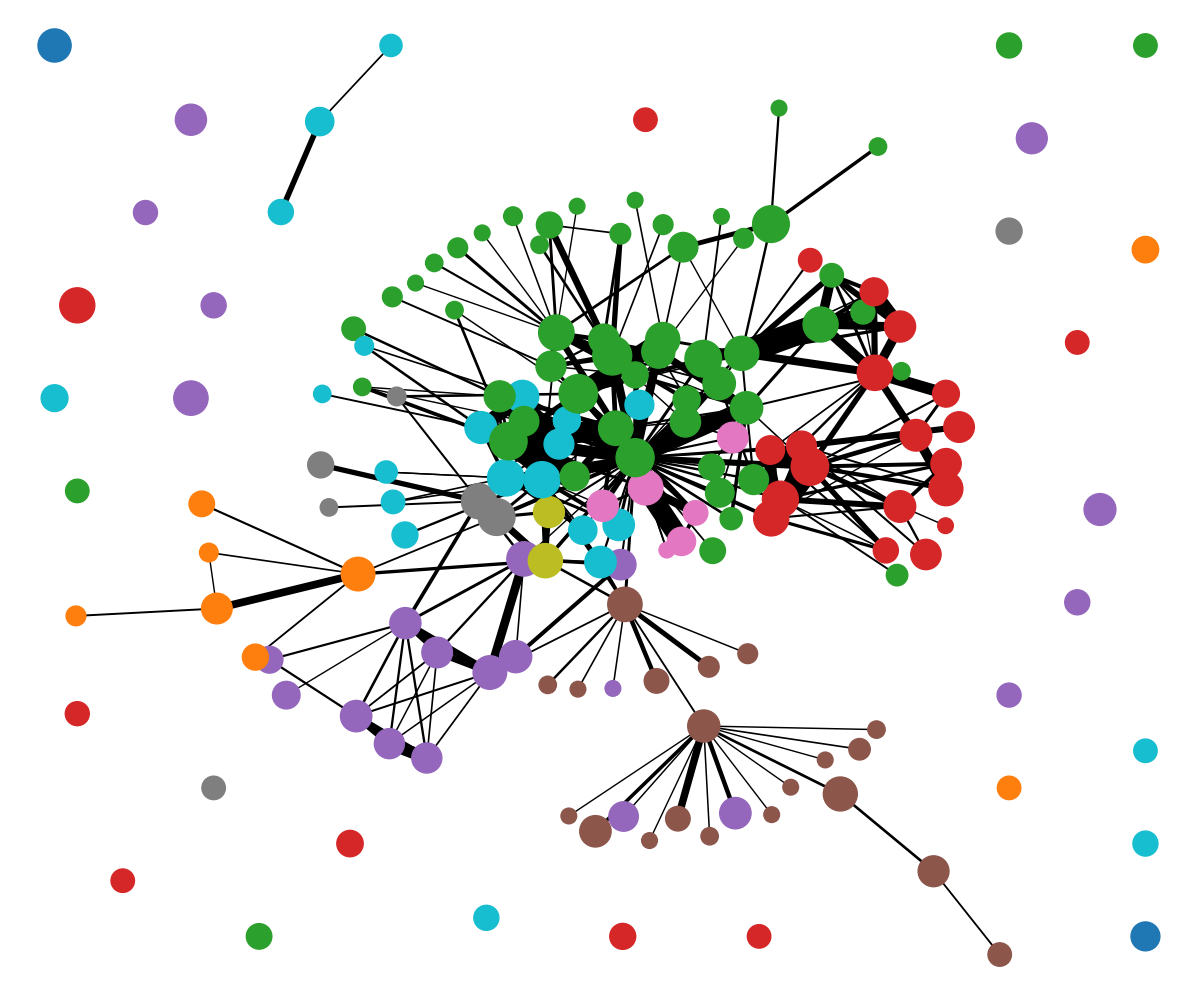} 
\caption{GAT predictions}
\end{subfigure}
\begin{subfigure}{0.49\columnwidth}
\includegraphics[width=\linewidth]{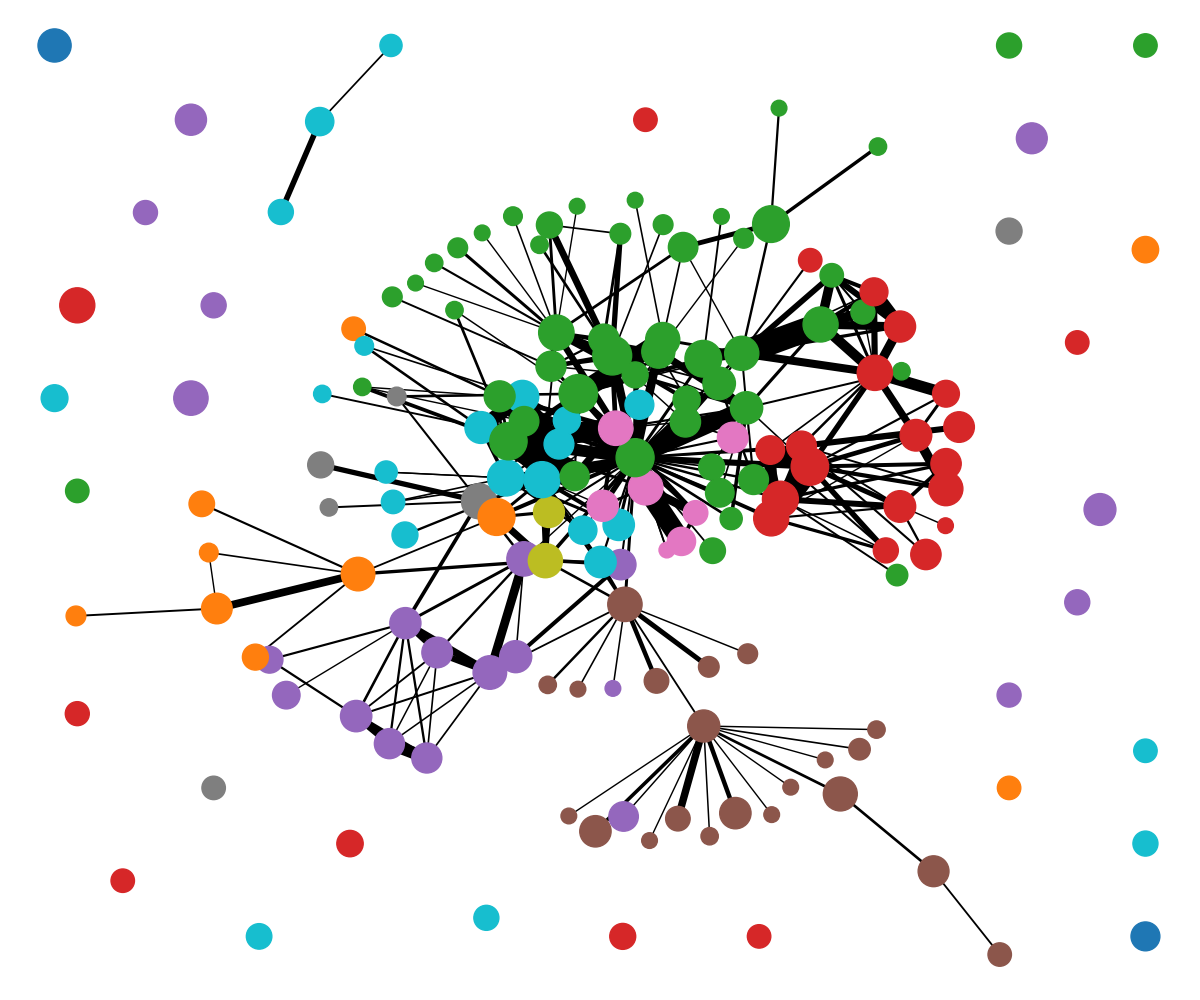} 
\caption{APPNP predictions}
\end{subfigure}

\caption{Deep Graph Mapper visualisation of the predictions of different node classification models. The top image colors each cluster according to its most frequent true label, similar to Figure \ref{fig:dgm-vis-wiki-cs}. The other plots have clusters colored according to the most frequent prediction of the appropriate model. Note that this can hide differences that do not change the majority prediction in a cluster. The specific parameters used are described in Appendix \ref{app:hyperparameters}.}
\label{fig:dgm-pred-vis}
\vskip -0.2in
\end{figure}

All DGM plots in Figures \ref{fig:dgm-vis} and \ref{fig:dgm-pred-vis} were created with the SDGM variant (which visualises the original graph structure), the unsupervised DGI lens, $t$-SNE reduction, 20 intervals, and the following additional parameters:

\begin{itemize}
    \item \textbf{Cora:} $\epsilon = 0.05$, min component size $8$.
    \item \textbf{CiteSeer:} $\epsilon = 0.03$, min component size $8$.
    \item \textbf{PubMed:} $\epsilon = 0.25$, min component size $12$.
    \item \textbf{Wiki-CS:} $\epsilon = 0.05$, min component size $10$.
\end{itemize}

\section{List of categories for each label}
\label{app:categories}

Table \ref{tab:categories} shows the list of categories used to construct each class. The listed categories have been selected by the sanitizer tool as aggregation targets, so each class consists of pages that were aggregated to one of the corresponding targets. Some subcategories of these targets that added unsuitable results were manually excluded.

\begin{table}[h]
\caption{List of aggregated Wikipedia categories used to construct each class.}
\label{tab:categories}
\vskip 0.15in
\begin{center}
\begin{small}
\begin{sc}
    \begin{tabular}{l|l}
\toprule
         \textbf{ID} & \textbf{Main categories}   \\
\midrule
         0 & Computational linguistics \\
         \midrule
         1 & Databases \\
         \midrule
         2 & \begin{tabular}[t]{@{}l@{}}
            Operating systems \\
            Operating systems technology
            \end{tabular} \\
         \midrule
         3 & Computer architecture \\
         \midrule
         4 & \begin{tabular}[t]{@{}l@{}}
            Computer security\\
            Computer network security\\
            Access control\\
            Data security\\
            Computational trust\\
            Computer security exploits
            \end{tabular} \\
            \midrule
         5 & Internet protocols \\
         \midrule
         6 & Computer file systems \\
         \midrule
         7 & Distributed computing architecture \\
         \midrule
         8 & \begin{tabular}[t]{@{}l@{}}
            Web technology\\
            Web software \\
            Web services
            \end{tabular}\\
         \midrule
         9 & \begin{tabular}[t]{@{}l@{}}
            Programming language topics \\
            Programming language theory \\
            Programming language concepts \\
            Programming language classification
        \end{tabular}\\
\bottomrule
\end{tabular}
\end{sc}
\end{small}
\end{center}
\vskip -0.1in
\end{table}

\newpage
\section{Deep Graph Mapper visualisation of model predictions}
\label{app:dgm-preds}

\end{document}